\documentclass{bmvc2k}
\usepackage[font=smaller]{caption}
\usepackage{wrapfig,booktabs}

\usepackage{ulem}
\usepackage{lipsum}

\title{Intrinsic Decomposition of Document Images In-the-Wild}

\addauthor{Sagnik Das}{sadas@cs.stonybrook.edu}{1}
\addauthor{Hassan Ahmed Sial}{hasial@cvc.uab.es}{2}
\addauthor{Ke Ma}{kemma@cs.stonybrook.edu}{1}
\addauthor{Ramon Baldrich}{ramon@cvc.uab.es}{2}
\addauthor{Maria Vanrell}{maria.vanrell@uab.cat}{2}
\addauthor{Dimitris Samaras}{samaras@cs.stonybrook.edu}{1}
\addinstitution{Computer Vision Lab\\
 Stony Brook University\\
 New York, USA
}
\addinstitution{Computer Vision Center\\
 Universitat Aut\`onoma de Barcelona\\
 Barcelona, Spain
}
\runninghead{Das, Sial, Ma, Baldrich, Vanrell, Samaras}{IDDocIIW}

\def\etal{\emph{et al}\bmvaOneDot}

\begin{document}
\maketitle
\begin{abstract}
Automatic document content processing is affected by artifacts caused by the shape of the paper, non-uniform and diverse color of lighting conditions. Fully-supervised methods on real data are impossible due to the large amount of data needed. Hence, the current state of the art deep learning models are trained on fully or partially synthetic images. However, document shadow or shading removal results still suffer because: (a) prior methods rely on uniformity of local color statistics, which limit their application on real-scenarios with complex document shapes and textures and; (b) synthetic or hybrid datasets with non-realistic, simulated lighting conditions are used to train the models.  In this paper we tackle these problems with our two main contributions. First, a physically constrained learning-based method that directly estimates document reflectance based on intrinsic image formation which generalizes to challenging illumination conditions. Second, a new dataset that clearly  improves previous synthetic ones, by adding a large range of realistic shading and diverse multi-illuminant conditions, uniquely customized to deal with documents in-the-wild.
The proposed architecture works in two steps. First, a white balancing module neutralizes the color of the illumination on the input image. Based on the proposed multi-illuminant dataset we achieve a good white-balancing in really difficult conditions. Second, the shading separation module accurately disentangles the shading and paper material in a self-supervised manner where only the synthetic texture is used as a weak training signal (obviating the need for very costly ground truth with disentangled  versions of shading and reflectance). The proposed approach leads to significant generalization of document reflectance estimation in real scenes with challenging illumination. We extensively evaluate on the real benchmark datasets available for intrinsic image decomposition and document shadow removal tasks. Our reflectance estimation scheme, when used as a pre-processing step of an OCR pipeline, shows a 26\% improvement of character error rate (CER), thus, proving the practical applicability. The data and code will be available at: \hyperlink{DocIIW}{https://github.com/cvlab-stonybrook/DocIIW}.
\end{abstract}
\vspace{8mm}
\section{Introduction}
\begin{figure}[!h]
    \centering
    \vspace{-2mm}
    \includegraphics[width=0.7\textwidth]{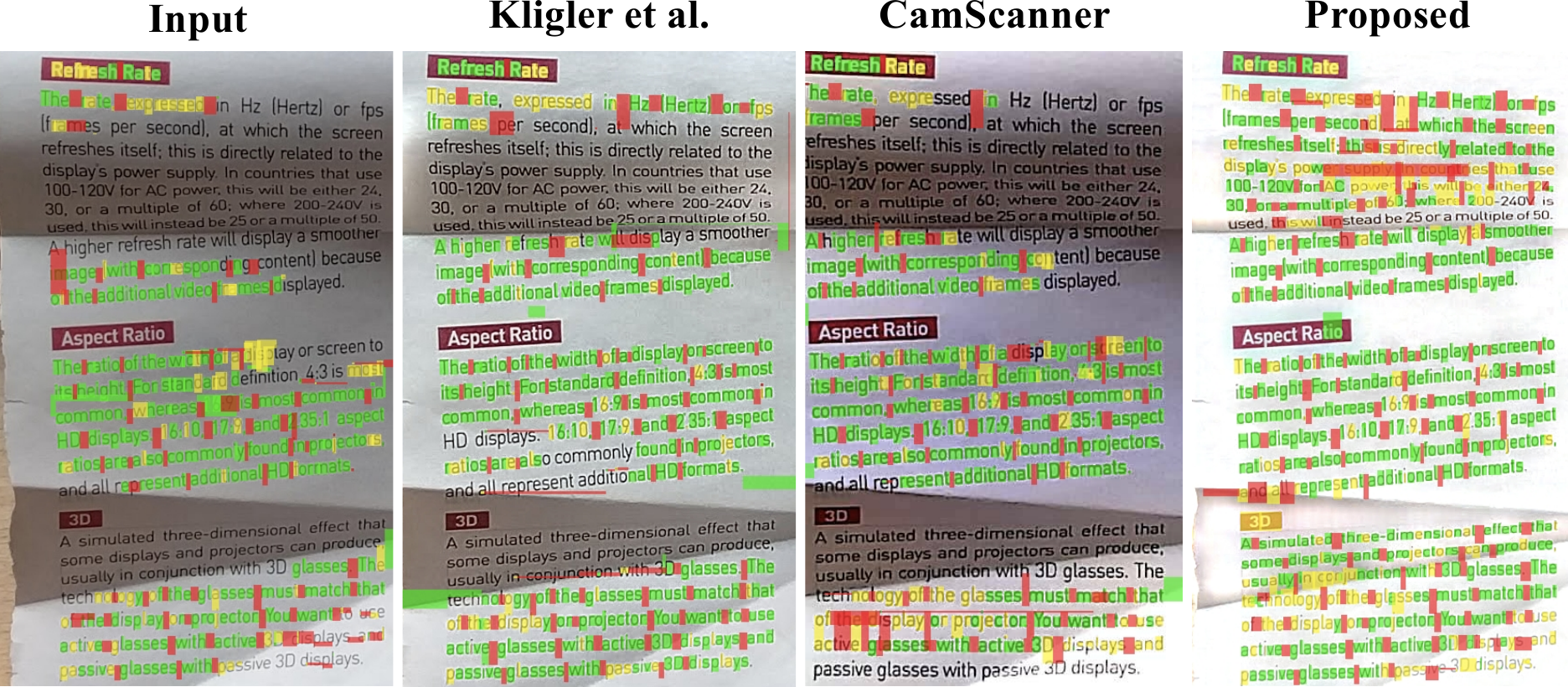}
    \vspace{3mm}
    \captionsetup{font=footnotesize}
    \caption{OCR accuracy (by Tesseract~\cite{nayak2014odia}) comparison on document images pre-processed with our proposed  method vs. Kligler \etal (state-of-the-art document shadow removal method~\cite{kligler2018document} ) vs. the commercially available document capturing application CamScanner.} \vspace{-2mm}
    \label{fig:teaser}
\end{figure}

\vspace{-2mm}
Paper documents contain and provide valuable information that is essential in our everyday life. By digitizing the documents, we can archive, retrieve, and share  contents with utmost convenience. With the increasing ubiquity of mobile camera devices, nowadays, capturing document images is the most common way of digitizing them. Such a document image is only useful if it can be used for further processing, such as text extraction and content analysis. Therefore, it is desirable to capture the image so that most of the document content is preserved. However, casual document images captured in-the-wild often contain substantial distortions due to non-uniform illumination conditions from multiple colored lights resulting in diverse shading and shadow effects. These physical artifacts hurt the accuracy and reliability of automatic information extraction methods. As an example, part of the text in the leftmost image of figure ~\ref{fig:teaser} remains undetected due to shadows and shading.
Recent deep learning image-to-image translation methods~\cite{isola2017image} are widely applicable but require large sets of training images to perform well in realistic scenarios. Enforcing additional domain-specific physical ground-truth helps to improve the generalization performance of such models~\cite{Das_2019_ICCV, sun2019single}. Alas, acquiring such ground-truth is often very costly. To model document illumination correction using deep learning, one would need a large document dataset with images, reflectance and shading captured in a large number of real scenes with various illumination conditions. It is very time-consuming and costly to collect such a dataset. In this paper, we propose a weakly-supervised method to separate reflectance and shading, and train the network by using physical constraints of image formation. Moreover, to facilitate the training  we  carefully curate a multi-illumination dataset, Doc3DShade, for document images in-the-wild that consists of realistic illumination and can be augmented to a much larger scale using rendering engines~\cite{blender_2018}.  

Prior document shadow and shading removal methods~\cite{bako2016removing, wang2019effective} mostly rely on local color statistics, thus imposing strong assumptions such as uniform background color of the document. Therefore, the application of these methods is generally limited to documents of a particular type with minimal colors, or graphics. Moreover, these methods assume the paper shape is almost flat~\cite{jung2018water, kligler2018document}, which also makes these approaches inapplicable to in-the-wild images of warped documents under challenging illumination conditions. To demonstrate the inability of existing methods in handling non-uniform illumination on complex paper shapes, we compare Optical Character Recognition (OCR) results in figure \ref{fig:teaser}. We compute OCR after pre-processing the input image (figure \ref{fig:teaser}) with the method of Kligler \etal.~\cite{kligler2018document} (current state-of-the-art for document shadow removal), a commercially available document processing application, and our proposed reflectance estimation scheme. Our method shows a significantly higher number of detected characters proving the effectiveness of the intrinsic image-based modeling of document images over traditional approaches. 

We address the aforementioned problems with a unified architecture based on two learning-based modules, WBNet and SMTNet, trained in a self-supervised manner on a new multi-illuminant dataset, Doc3DShade, built on top of the public Doc3D \cite{Das_2019_ICCV} dataset. We formulate the problem of document reflectance estimation based on the physics of image formation under the Lambertian assumption, i.e., an image $I$, is a composition of the shading $S$ and reflectance $R$, $I=R \bigotimes S$, where $\bigotimes$ denotes the Hadamard product.
Images in the Doc3Dshade are created by following this assumption. In Doc3DShade we capture realistic shading $MS$ of a deformed non-textured document under a large range of realistic colors, and diverse multi-illuminant conditions. Note that $MS$ contains both the shading and the color of the paper material. We render real document textures $T$ in Blender~\cite{blender_2018} and create $I$ by combining $I=T\bigotimes MS$. Our formulation is similar by considering $R=T \bigotimes M$. Since the shading component is physically captured in controlled environments, these images are clearly superior in terms of physical illumination effects than the existing Doc3D~\cite{Das_2019_ICCV} images. We have generated 90,000 images in Doc3DShade which effectively double the size of Doc3D.

The WBNet and SMTNet address white-balancing and shading removal respectively. WBNet inputs an RGB image  and estimates a white-balanced image where  the illuminant color is removed. We can train WBNet with  Doc3DShade which contains a white-balanced image for every input. In the second module, SMTNet takes the white-balanced image and regresses a per-pixel shading image and paper material image. The use of the white-balanced image allows us to train SMTNet in a self-supervised manner using a physical constraint, the chromatic consistency of the white-balanced and the reflectance image. Chromatic consistency states that the white-balanced output's chromaticity is same as true reflectance of the paper and the texture.
We thus, eliminate the costly need to physically capture explicit  shading and paper reflectance ground truth. Our method shows strong results on challenging illumination conditions and generalizes across different document types. The practical significance of the proposed method is demonstrated by a 21\% decrease in OCR error rate when our shading removal is used as pre-processing for OCR.

In summary our contributions are:
First, a unified learning-based  architecture that directly estimates document reflectance based on an intrinsic image formation model, that generalizes across challenging illumination conditions for different types of documents as well as for warped documents.

\noindent Second, we created Doc3Dshade, a new dataset that clearly improves previous synthetic or hybrid ones, by physically acquiring a  range of realistic color diverse multi-illuminant conditions and using them to synthesize a multitude of complexly illuminated document images. This dataset is uniquely customized to deal with documents in-the-wild.

\vspace{-5mm}
\section{Related Work}
\vspace{-2mm}
\noindent\textbf{Color constancy(CC)} is the visual phenomenon of perceiving the true  object color,  independently of ambient lighting. Computational color constancy aims to estimate the color of the image scene, used for white-balancing, i.e. a corrected version of the image under an achromatic illuminant. It is a standard pre-processing step in document analysis tasks \cite{kligler2018document}.
Both  statistics based methods\cite{van2007edge,gijsenij2010color} and deep learning based methods \cite{bianco2015color,lou2015color,hu2017fc4} have used a uniform illumination model to solve the CC task. More recent methods focus on the multi-illuminant CC problem, working on pairs of images taken under different illuminants \cite{hui2016white, hui2019learning} or using two-branch networks to describe local illuminants \cite{shi2016deep}. Our white-balancing network (WBNet) is a single image color constancy model targeted to document images.


\noindent\textbf{Intrinsic image decomposition} is an inverse optics formulation that can be helpful in removing shading and shadows effects in document images.
 Early work in this domain \cite{barrow1978recovering,horn1974determining,Weiss2001,Tappen2006,Funt1992} is dominated by Retinex theory \cite{Land1971}. 
Later work added more physical priors on reflectance and shading \cite{Gehler2011,LShen2013,Barron2015,bell14intrinsic}.
Recent supervised deep learning approaches \cite{Narihira2015,Kim2016UnifiedDP} have used end-to-end regression models to predict all the intrinsic modalities. More recently constrains such as physical intrinsic losses \cite{Fan2018RevisitingDI,Baslamisli18,Sial:20} and human judgments \cite{Zhou2015LearningDR,Nestmeyer2017ReflectanceAF} were imposed on the network to solve this problem. Unsupervised and semi-supervised models generally use multiple illumination varying sequences of the same scene to learn this decomposition \cite{Ma2018SingleII,LVvG18,janner2017self,ZhenggiCVPR2018Learning}. We only use single images and manage to train our decomposition framework using synthetic texture as a weak-supervision signal.

\noindent\textbf{Shadow and shading removal} of lighting variations 
has been widely studied. Most  shadow removal methods, formulated for natural outdoor images, do not perform well on document images because of the specific characteristics of printed text. Earlier methods used an intrinsic approach to remove shading effects from the images \cite{brown2006geometric,zhang2007removing},    document colour homogeneity \cite{bako2016removing},  neighbouring pixel values \cite{wang2019effective}, or combined visibility detection with existing state-of-the-art methods \cite{kligler2018document} or finally,  applied diffusion equations on pixel intensity \cite{jung2018water}.

\noindent\textbf{Self-supervised learning} alleviates the need of large-scale labeled data. It exploits unlabeled image/video attributes, automatically generating pseudo labels for training. Self-supervised learning tasks  include image colorization (an RGB image as the target and its grayscale version as the input)~\cite{zhang2016colorful}, image jigsaw puzzle (patches from one image as the input and their spatial relation as the target)~\cite{doersch2015unsupervised}, temporal order verification (frames from one video as the input and their temporal relation as the target)~\cite{wei2018learning}.  Drawing ideas from these prior methods, we train the shading decoder of SMTNet to produce a shading image consistent with the shading image estimated from the material decoder output.

\vspace{-4mm}
\section{Training Dataset: Doc3DShade}
\vspace{-2mm}
We collected a large-scale documents dataset, Doc3DShade, which combines diverse, realistic illumination scenarios with natural paper textures.
Doc3DShade increases the physical correctness of Doc3D, a recent document dataset~\cite{Das_2019_ICCV}. The contributions  of Doc3DShade are two-fold; first, it captures physically accurate and realistic shading under complex illumination conditions, which is impossible to obtain using rendering  engines (which use approximate illumination models to simulate light transport).  Second, it contains various paper materials with different reflectance properties whereas a synthetically rendered dataset like Doc3D uses purely diffuse material to render images. Doc3DShade is thus physically more accurate and can be used to model scene illumination   in a physically grounded way. 
\begin{figure}
    \centering
    \includegraphics[width=0.8\textwidth]{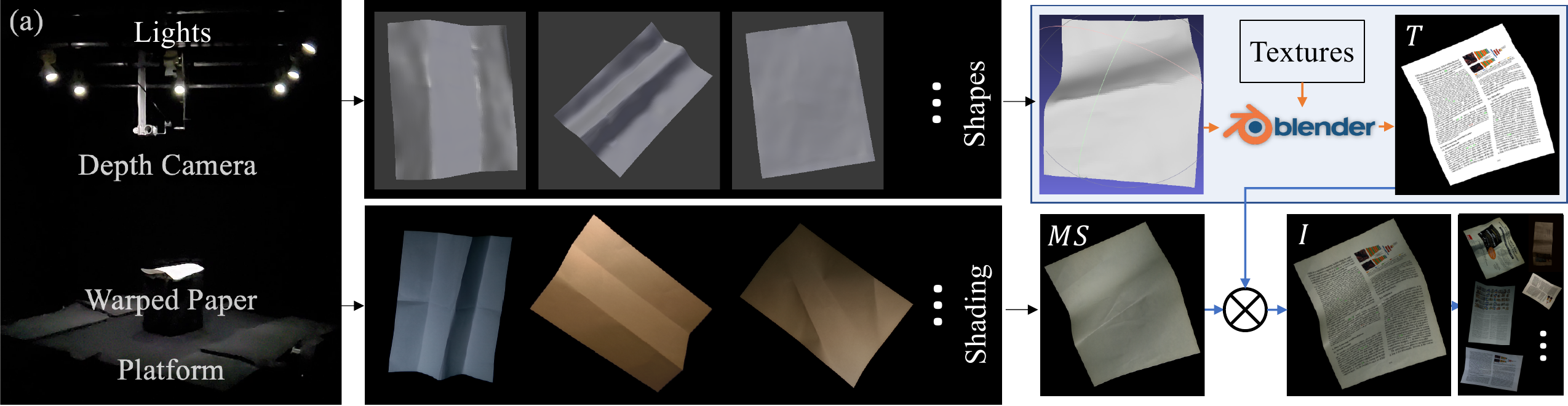}
    \vspace{2mm}
    \captionsetup{font=footnotesize}
    \caption{Data Creation Pipeline: (a) Shows the hardware setup. The captured shapes are textured in Blender and combined with the captured shading image by Hadamard product ($\bigotimes$) to create $I$. The `orange' and `blue' arrows denote the rendering and the combining.}  
    \label{fig:data_pipe}
     \vspace{-6mm}
\end{figure}
 \noindent\textbf{Capturing 3D Shape and Shading:} Our capturing setup (figure \ref{fig:data_pipe}(a)) consists of a rotatory platform, $8$ directional lights and a depth camera. More details are  in the supplementary material.
  To capture $3D$ shapes of documents, we randomly deformed textureless papers and placed them on the rotatory platform. The platform is randomly rotated and each document is captured with a random combination of single and multiple color directional lights. Light chromaticity was constrained to be around the Planckian locus \cite{colorscience07} to simulate natural lighting conditions. We also captured each document under white lights to use it later as ground-truth for white balancing. To include various material properties in the shading images, we have used $9$ diffuse paper materials that are used in various types of documents such as magazines, newspapers and printed papers, etc. In summary, we obtain the following data for each warped paper: a 3D point cloud and multiple images with shading under single and multiple lights of different color temperatures. An illustration of the captured shapes and corresponding shading images are shown in Fig. \ref{fig:data_pipe}. Note that the captured shading ($MS$) map in this setup is a combined form of the paper material ($M$) and the shading ($S$) component.
 
\noindent\textbf{Image Rendering:} We create the 3D mesh for each point cloud following~\cite{Haefner2018CVPR}. Each mesh is textured with a random document image and rendered with diffuse white material in Blender~\cite{blender_2018} (Fig.~\ref{fig:data_pipe}). In total, we have used $\sim 5000$ textures collected from various documents such as magazines, books, flyers, etc. Each texture image is combined with a randomly selected single light shading image of the same mesh. To create more variability, we uniformly sample and linearly combine two single light shading images to simulate a fake multi-light shading image: $I=T\bigotimes(a.MS_1 + (1-a).MS_2)$, with $a \in [0,1]$. Additionally, for each image, we also render the white balanced image by combining the synthetic texture image and the shading image captured under white light.  We have created 90K images  with  diffuse texture and  white-balance images as ground-truths, for training and testing.
\vspace{-3mm}
\section{Proposed Method}
\vspace{-2mm}
Our reflectance estimation framework for document images captured under non-uniform lighting in a real-world scenario follows a two-step approach for illuminant and shading correction and leverages the physical properties of the image formation model. In the first step, we estimate a white-balanced image to neutralize the color of the scene illuminants. In the second step, we disentangle shading, texture, and material of the document, which allows us to obtain the shading-free image. 
\vspace{-3mm}
\subsection{Image Formation Model.} 
\vspace{-1mm}
We assume the scene is Lambertian, and illuminated by $n$ light sources $l$. The color of each light source  $l_i$  is represented as a three channel vector$(l_i^{r}, l_i^{g}, l_i^{b})$.  $\lambda_i$  is the shading induced by the $i$-th light. Under the Lambertian assumption with no inter-reflections, image intensity $I$ at pixel $p$ can be modeled as:
\vspace{-3mm}
\begin{equation}
    I^{c}(p)=R^{c}(p) \sum_{i}\lambda_i(p)l_i^{c}
    \vspace{-2mm}
    \label{eq:imgmodel}
\end{equation}
where $R^{c}(p)$ is the reflectance at pixel $p$ and $c \in \{r,g,b\}$ denotes the color channel. In the  case of documents we can assume the document reflectance is a combination of a \textit{diffuse texture} component, i.e. the content of the documents and a \textit{material} component, i.e. the color of the paper. Therefore we can rewrite the image formation model (Eq. \ref{eq:imgmodel}) as:
\begin{equation}
    I^{c}(p)=M^{c}(p)T^{c}(p) \sum_{i}\lambda_i(p)l_i^{c} 
    \label{eq:docimgmodel}
        \vspace{-2mm}
\end{equation}
with $M$ and $T$ the reflectance  of \textit{material} and \textit{texture} respectively. The shading-free image we want to recover is the product of $M$ and $T$. $T$ is available as ground-truth in Doc3DShade, but $M$ is combined with the shading term thus not explicitly available as  ground-truth.
\vspace{-3mm}
\subsection{Problem Formulation.}
\vspace{-1mm}
Considering the image model given in Eq. \ref{eq:docimgmodel} we will first eliminate illuminant color effects through white-balancing and then disentangle material and texture.

We estimate a white-balanced image version $I_{wb}$ of the input image. It has the same reflectance properties as the original images, given by Eq. \ref{eq:docimgmodel}, but under multiple achromatic lights, i.e. $l_i=(\eta_i, \eta_i, \eta_i)$. We follow similar formulation used by Hui et al. in \cite{iccp16wp}: 

\vspace{-1mm}
\begin{equation}
    I_{wb}^{c}(p)=M^{c}(p)T^{c}(p) \sum_{i}\lambda_i(p)\eta_i
    \label{eq:docwbmodel}
\end{equation}
To obtain the white-balanced image, $I_{wb}$, we follow the formulation of \cite{iccp16wp}, where $I_{wb}$ is estimated in terms of a per-pixel white-balance kernel $WB(p)\in \mathcal{R}^3$  which corrects the color of the incident illumination at every single pixel, $p$, thus:
\vspace{-1mm}
\begin{equation}
     I_{wb}^{c}(p)= WB^{c}(p)I^{c}(p)
     \label{eq:wbk}
\end{equation}
Since the shading $\lambda_i$ is invariant to the light color,  per pixel chromaticity\footnote{Chromaticity of a pixel is given as: $C_r=r/(r+g+b),C_g=g/(r+g+b),C_b=b/(r+g+b)$} for  $I_{wb}$ is:
\vspace{-1mm}
\begin{equation}
C_{wb}^{c}(p)=\frac{R^{c}(p) \sum_{i}\lambda_i(p)\eta_i}{\sum_{c}R^{c}(p) \sum_{i}\lambda_i(p)\eta_i} = \frac{R^{c}(p)}{\sum_{c}R^{c}(p)}
\label{eq:chromwb}
\end{equation}
$C_{wb}^{c}(p)$ is identical with the chromaticity of the Reflectance component $R^{c}(p)$ which is the physical constraint used in our self-supervised loss. Eq. \ref{eq:chromwb} can  also be expressed in terms of the texture and material components as following:
\vspace{-1mm}
\begin{equation}
    C_{R}^{c}(p)=\frac{R^{c}(p)}{\sum_{c}R^{c}(p)} =  \frac{M^{c}(p)T^{c}(p)}{\sum_c M^{c}(p)T^{c}(p)}
    \label{eq:chromref}
\end{equation}

\begin{figure}[!h]
    \centering
    \includegraphics[width=0.8\textwidth,height=1.6in]{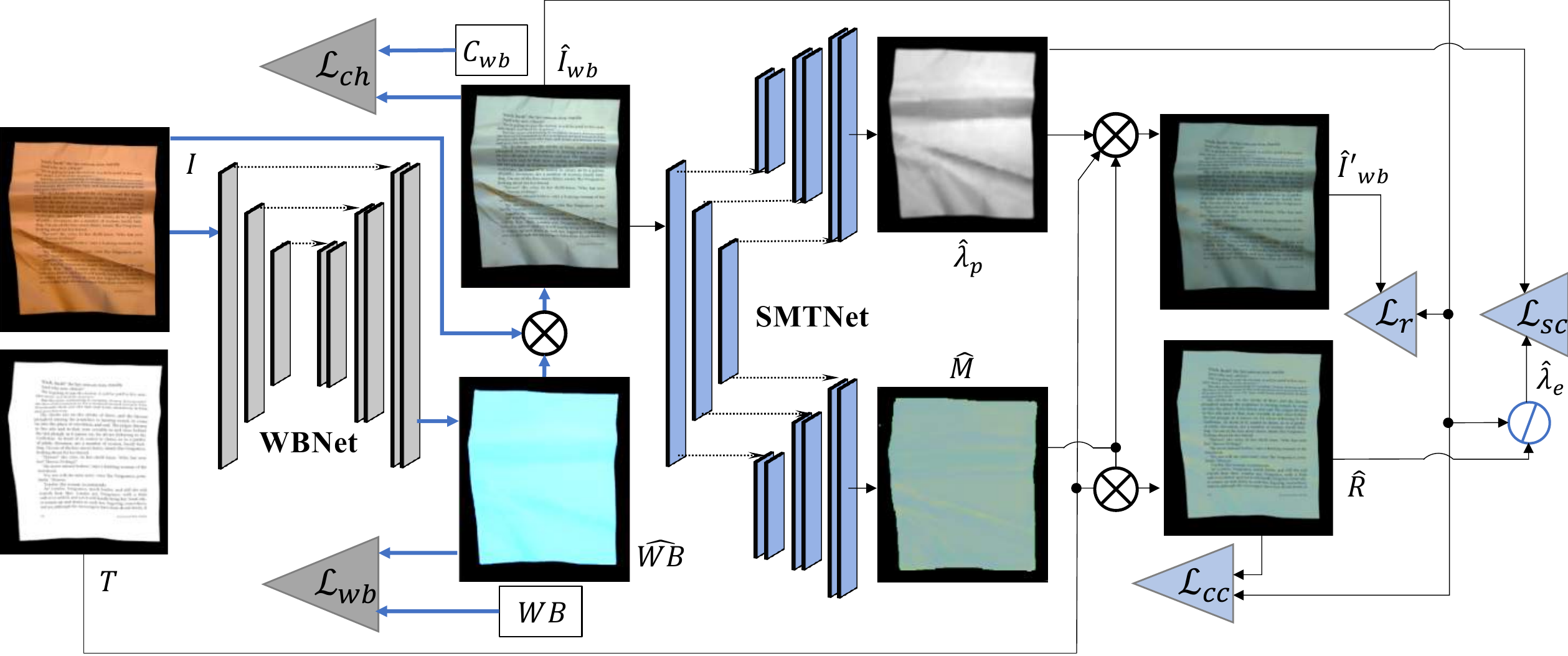}
    \vspace{4mm}
    \captionsetup{font=footnotesize}
    \caption{Proposed framework: The WBNet takes RGB image, $I$ as input and produces the white-balance kernel ($\hat{WB}$). The white-balanced image, $\hat{I}_{wb}$ is then forwarded to the SMTNet which regresses the material ($\hat{M}$) and the shading, $\hat{\lambda}_{p}$. The $\bigotimes$ denote the Hadamard product, $(/)$ denote division and the triangles denote the loss functions.}
    \vspace{-5mm}
    \label{fig:WBNet_SMTNet}
\end{figure}

In what follows, we are going to use the derived physical constraints, $C_{wb}^{c}=C_{R}^{c}$, to estimate the intrinsic components of a document image given a single image $I^{c}$. Our approach is based on using two sub-networks- WBNet and SMTNet (Fig. \ref{fig:WBNet_SMTNet}). The first network WBNet estimates the white-balance kernel, $WB^{c}(p)$, whose chromaticity, $C_{wb}$, is used in the subsequent network, SMTNet, that in turn will estimate the shading, $\lambda_i(p)$ and the material, $M^{c}(p)$ in a self-supervised fashion. We choose to employ self-supervised training since separate ground-truths are not available for $\lambda_i(p)$ and $M^{c}(p)$. 


\vspace{-2mm}
\subsection{WBNet: White-balance Kernel Estimation.}
\vspace{-1mm}
The first sub-network is designed to estimate the per-pixel white-balance kernel $WB$. Since the number of illuminants, their colors, and the individual shading terms are unknown, estimating the white-balanced image becomes an ill-posed problem given a single image. We treat this task as an image-to-image translation problem. Given an input color image, $I\in \mathcal{R}^{h\times w \times 3}$ the white-balance kernel, $WB \in \mathcal{R}^{h\times w \times 3}$ is estimated using a UNet \cite{ronneberger2015u} style encoder-decoder architecture with skip connections. The predicted white-balanced image, $\hat{I}_{wb}$  is then estimated using  Eq. \ref{eq:wbk}, by Hadamard product of $\hat{WB}$ and $I$. Note that although it is possible to directly estimate the white-balanced image $\hat{I}_{wb}$ from $I$, convolutional encoder-decoder architectures cannot preserve sharp details such as text. On the other hand, $WB$ is a smooth per-pixel map that is much easier to learn in reality.

\noindent\textbf{Loss Function:} To train the WBNet we essentially use two loss terms, one on the predicted white-balance kernel ($\mathcal{L}_{wb}$) and another on the chromaticity of the estimated white-balanced image ($\mathcal{L}_{ch}$). As seen from Eq. \ref{eq:chromwb}, chromaticity is a shading invariant term, as is also the white-balance kernel, $WB$. Thus, it is intuitive to learn $WB$ using a loss that is shading invariant. Therefore, we prefer to use the $\mathcal{L}_{ch}$ instead of a standard reconstruction loss on the predicted white-balanced image, $\hat{I}_{wb}$. We validate our choice in an ablation study in the supplementary material.

Additionally, we force  intensity at each pixel to be preserved after white-balancing, which acts as a constraint on the predicted image. The combined loss is:
\vspace{-2mm}
\begin{equation}
\mathcal{L}_{wbn}=\mathcal{L}_{wb}(\hat{WB}, WB)+\alpha_1 \mathcal{L}_{ch}(\hat{C}_{wb}, C_{wb})+\alpha_2 \mathcal{L}_{int}(\hat{In}_{wb}, In)
\label{eq:wbkloss}
\end{equation}
where $\hat{WB}$, $\hat{C}_{wb}$, $\hat{In}_{wb}$ are the predicted white-balance kernel, chromaticity and intensity image of the predicted  white-balanced image respectively. Corresponding ground-truths are available in our training dataset. Per-pixel intensity image is  $\hat{In}_{wb} (p)= \sum_{c}\hat{I}_{wb}^c (p)$. The $\alpha$'s are the weights associated with each loss term. We use $L_1$ distance for each loss term. For  $\mathcal{L}_{wb}$ and $\mathcal{L}_{ch}$ losses we use a mask to avoid pixels where  ground-truth pixel values are zero. 
\vspace{-2mm}
\subsection{SMTNet: Separating Material, Texture and Shading.}
\vspace{-1mm}
Given the estimated white-balanced image $\hat{I}_{wb}$ the second module, SMTNet estimates the material $\hat{M} \in \mathcal{R}^{h\times w \times 3}$ and shading $\hat{\lambda}=\sum_i\lambda_i \in \mathcal{R}^{h\times w \times 1}$. The structure of the network is illustrated in Fig. \ref{fig:WBNet_SMTNet}. The network consists of one encoder and two identical decoder branches, MNet and SNet. MNet and SNet regress the $\hat{M}$ and $\hat{\lambda}$ from the input image respectively. Ideally, it is possible to separately learn  $\hat{M}$ and $\hat{\lambda}$ in a supervised manner if the corresponding ground-truths are available. But captured shading ($MS$) in our dataset is a combined representation of $M . \lambda$ which is further combined with the diffuse textures $T$ to generate our training images $I=M . T . \lambda$ (Eq. \ref{eq:docwbmodel}). Therefore, we resort to a self-supervised approach to train SMTNet for disentangling $M$ and $\lambda$ given $I$, assuming $T$ is available as ground-truth. 

\noindent\textbf{MNet}: Given the predicted material image, $\hat{M}$ from the MNet branch, we can estimate the reflectance image as $\hat{R}=\hat{M}.T$ and estimate the shading image as $\hat{\lambda}_e=I/(\hat{M}.T)$. To train this branch, we use the consistency property of Eq. \ref{eq:chromwb} and Eq. \ref{eq:chromref}. If the predicted material $\hat{M}$, is correct the chromaticity of the $\hat{R}$ and chromaticity of the input white-balanced image should be the same. We use this chromatic consistency loss  to train the MNet branch.  

\noindent\textbf{SNet}: The other branch, SNet predicts the shading image $\hat{\lambda}_p$. To ensure the consistency of predicted material and predicted shading, $\hat{\lambda}_p$ should be equal to the estimated shading from the MNet branch, $\hat{\lambda}_e$. We use this  shading consistency loss to train the SNet branch.

The SMTNet is a modified UNet architecture with a single encoder and two separate decoders. Decoders share encoded features but use different skip connections to the encoder.  

\noindent\textbf{Loss Function:} The two primary loss functions used to train the SMTNet are the chromatic consistency ($\mathcal{L}_{cc}$) and the shading consistency ($\mathcal{L}_{sc}$) loss. Additionally, we use the reconstruction loss to ensure the predicted material image, $\hat{M}$ and shading image, $\hat{\lambda}_p$ reconstruct the input image when combined with the input diffuse texture, $T$. We also add  smoothness constraints on the predicted material and shading. Specifically, the $L_1$ norm of the gradients for the material, $||\nabla\hat{M}||$ and $L_1$ norm of the second order gradients for the shading, $||\nabla^2\hat{\lambda}_p||$ are added as regularizers. Whereas the first term ensures piecewise smoothness of the $\hat{M}$, the second term ensures a smoothly changing shading map. The combined loss function is:
\vspace{-1mm}
\begin{equation}
\mathcal{L}_{smt}=\mathcal{L}_{cc}(\hat{C}_{wb}, \hat{C}_{R})+\beta_1 \mathcal{L}_{sc}(\hat{\lambda}_p, \hat{\lambda}_e)+\beta_2 \mathcal{L}_{r}(\hat{I'}_{wb}, \hat{I}_{wb}) + \beta_3 ||\nabla^2\hat{\lambda}_p|| +\beta_4 ||\nabla\hat{M}||
\label{eq:smtloss}
\end{equation}
Where $\hat{C}_{wb}$ and  $\hat{C}_{R}$ are the chromaticities of input white-balanced image and estimated reflectance image. The $\hat{\lambda}_p$ and $\hat{\lambda}_e$ are the predicted and estimated shading. $\hat{I'}_{wb}$, $\hat{I}_{wb}$ are the reconstructed and input white-balanced image. The $\hat{I'}_{wb}$ is estimated as: $\hat{M}.T.\hat{\lambda}_p$, the hadamard product of predicted material image, shading image and input diffuse texture. The $\beta$'s are the weights associated with each loss term. For each loss term we use the L1 loss.

Further, we also experiment with adversarial loss on the predicted shading map. The adversarial loss turns out to be helpful to remove texture artifacts from the predicted shading map. We use the WGAN-GP loss \cite{gulrajani2017improved} with a PatchGAN \cite{isola2017image} critic by treating the estimated shading $\hat{\lambda}_e$ as the examples from the real distribution. The modified $\mathcal{L}_{smt}$ with the adversarial loss term is given as:
\begin{equation}
    \mathcal{L}_{smt\_g}=\mathcal{L}_{smt} + \beta_5 \mathcal{L}_{wgan}(\hat{\lambda}_p, \hat{\lambda}_e)
\end{equation}

\vspace{-4mm}
\section{Evaluation}
\vspace{-1mm}
 For the quantitative and qualitative evaluation of our approach, we have experimented with multiple datasets that are captured under varying illumination conditions. Our evaluation contains three comparative studies. First, we show how the proposed method behave in intrinsic decomposition of document images. Second, we use the proposed approach as a post-processing step in a document unwarping pipeline~\cite{Das_2019_ICCV} to show the practical applicability of our method in terms of OCR errors. Third, we show an extensive qualitative comparison with current state-of-the-art document shadow removal methods~\cite{bako2016removing, jung2018water, wang2019effective, kligler2018document}. From these experiments, we show a 26\% improvement in OCR when used as a pre-processing step and also show strong qualitative performance in intrinsic document image decomposition and shadow removal tasks.

\vspace{-1mm}
\subsection{Intrinsic Document Image Decomposition.}\label{subsec:intrinsic}  
\vspace{-1mm}
\begin{wraptable}[10]{r}{7cm}
\vspace{-3mm}
\centering
\resizebox{0.5\textwidth}{!}{\begin{tabular}{@{}ccccc@{}}
\toprule 
\textit{DewarpNet} & \multicolumn{2}{c}{\textbf{Image Quality}} & \multicolumn{2}{c}{\textbf{OCR Errors}}\\
\textit{results}          & MS-SSIM & LD & CER                 & WER                 \\ \midrule
with shading       & 0.4692 &  8.98  & 0.3136  & 0.4010 \\
w/o shd~\cite{Das_2019_ICCV} & 0.4735 &  8.95  & 0.2692 & 0.3582 \\
\midrule
w/o shd (Ours)        & \textbf{0.4792}  &  8.74  & 0.2453  & 0.3325 \\ 
w/o shd+adv (Ours)        & 0.4778  &  \textbf{8.51}  & \textbf{0.2312}  & \textbf{0.3209} \\ \bottomrule
\end{tabular}}
\vspace{4mm}
 \captionsetup{font=footnotesize}\captionof{table}{Quantitative comparison of ~\cite{Das_2019_ICCV}'s unwarping quality on DocUNet benchmark dataset~\cite{ma2018docunet} when our proposed approach is applied as a pre-processing step before OCR. `w/o shd+adv' denote the model trained with adversarial loss. }
 \label{tab:dwarp}
\end{wraptable}
In this section we evaluate the performance of our method in decomposing intrinsic light components on two datasets e.g., DocUNet benchmark~\cite{ma2018docunet} and MIT Intrinsics~\cite{bell14intrinsic}. The DocUNet benchmark contains shading and shadow-free flatbed scanned version. Practically, scanned images are the best possible way to digitize a document and can be considered as the reflectance ground-truth. For this experiment, we apply our method on the benchmark images, then use the pre-computed unwarping maps from~\cite{Das_2019_ICCV} to unwarp each image which aligns each image with its scanned reflectance ground-truth. The quantitative results for image quality before and after applying our approach is reported in table \ref{tab:dwarp}, in terms of the the unwarping metrics: multi-scale structural similarity (MS-SSIM)~\cite{wang2003multiscale} and Local Distortion (LD)~\cite{you2017multiview}. The qualitative results are shown in figure \ref{fig:dwarp}. Improvement of the MS-SSIM and LD are limited since these metrics are more influenced by the quality of the unwarping. Qualitative images show significant improvement over shading removal applied in ~\cite{Das_2019_ICCV}. It is due to explicit modeling of the paper background and illumination, which enables our model to retain a consistent background color. Few additional qualitative results on real images captured under complex illumination are shown in figure \ref{fig:realcomp}. Intermediate output of WBNet in figure \ref{fig:realcomp} demonstrates good generalization to real scenes.

\noindent \begin{minipage}[t]{0.42\textwidth}
\centering
\vspace{1mm}
\includegraphics[width=\textwidth]{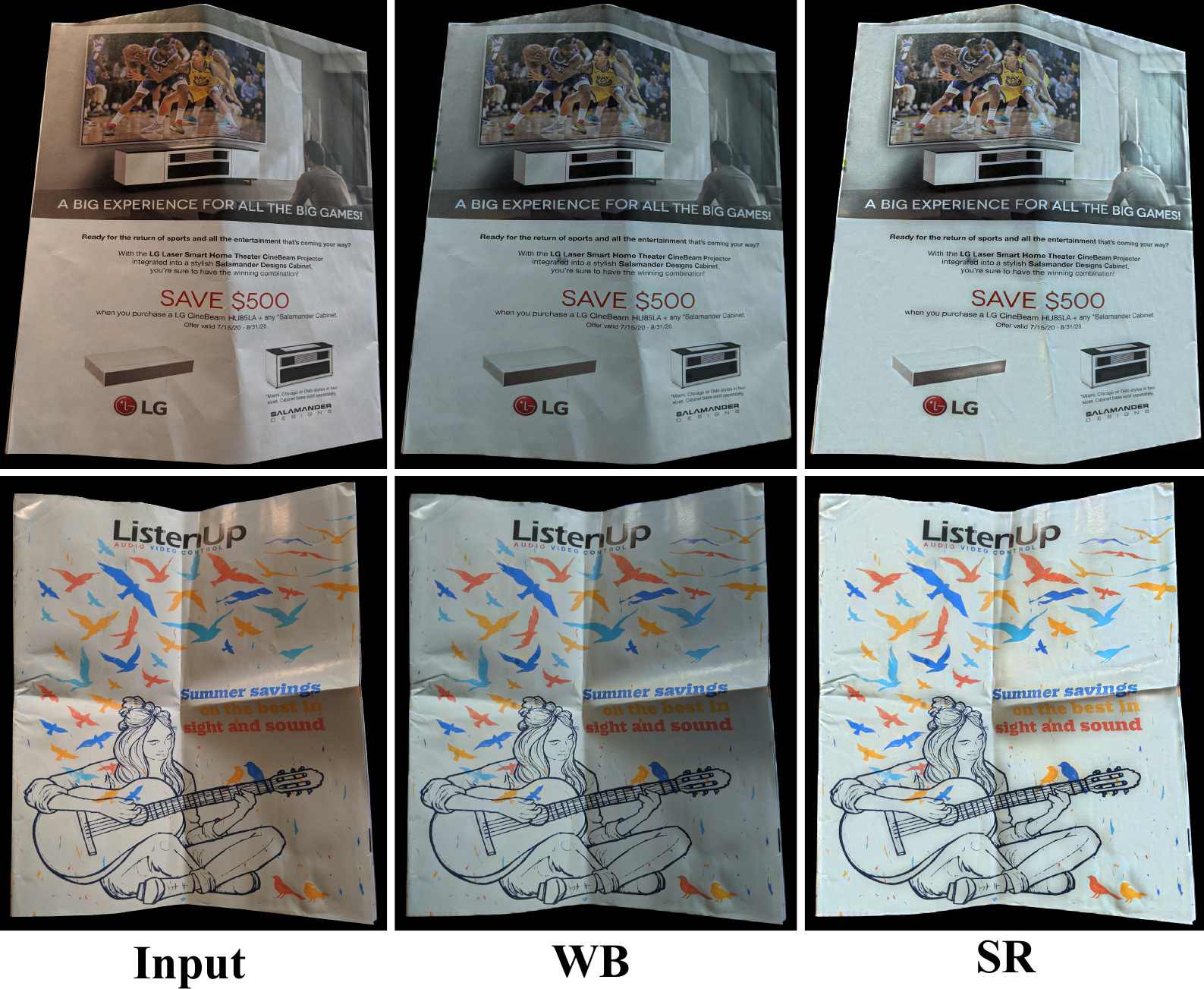}
\vspace{-1mm}
\captionsetup{font=footnotesize}
\captionof{figure}{Qualitative results on real images after applying white balancing (After WB) and shading removal (After Shd. Rem.). Input images are non-uniformly illuminated with two lights.}
\vspace{1.5mm}
\label{fig:realcomp}
\end{minipage}\hfill \begin{minipage}[t]{0.56\textwidth}
\vspace{1mm}
\includegraphics[width=\textwidth]{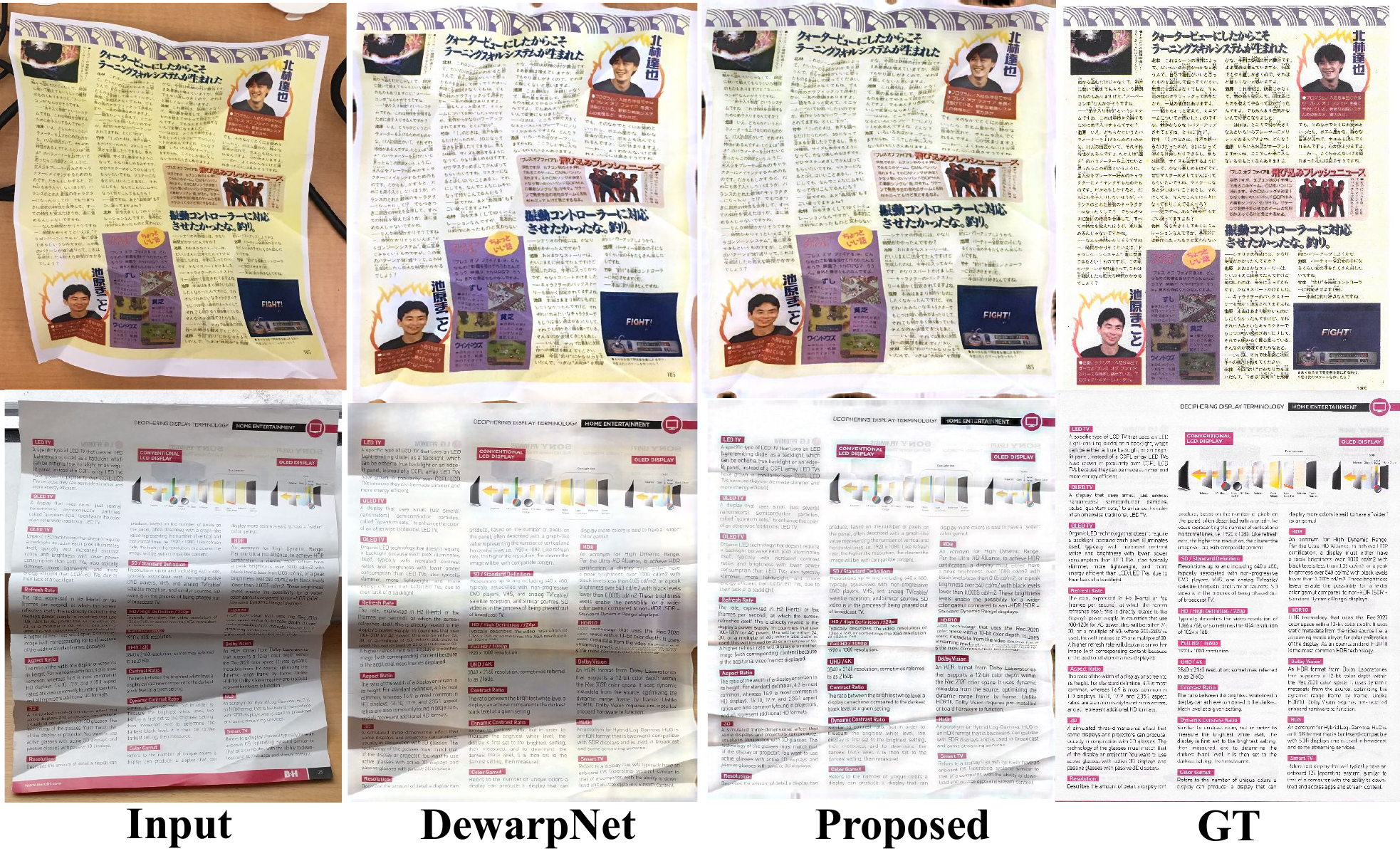}
\vspace{0.5mm}
\hfill
\captionsetup{font=footnotesize}
\captionof{figure}{Results on real-world images from \cite{ma2018docunet}. \cite{Das_2019_ICCV}'s shading removal method fails to retain the background since shading, illumination and document background is modeled as a single modality.}
\vspace{1.5mm}
\label{fig:dwarp}
\end{minipage}
Additionally, in figure \ref{fig:mitwetzu_comp}(a), we show that our method generalizes for 'paper-like' objects of  MIT-Intrinsics~\cite{bell14intrinsic}, e.g., paper and teabag objects without any fine-tuning. Interestingly, IIW results show that it fails to preserve the text on the 'teabag', which proves general intrinsic methods are probably not suitable for document images and calls for further research attention to specially design intrinsic image methods for documents.
\vspace{-3.5mm}
\subsection{Pre-processing for OCR.}\label{subsec:OCRevaluation}
\vspace{-1mm}
To demonstrate practical application of our approach, we compare OCR performance on the DocUNet benchmark images using word error rate (WER) and character error rate (CER) detailed  in~\cite{Das_2019_ICCV}. At first, the same unwarping step is applied as described in section \ref{subsec:OCRevaluation}. Then we perform OCR (Tesseract) on the DewarpNet OCR dataset before and after applying the shading removal and compare the proposed approach with the shading removal scheme presented in~\cite{Das_2019_ICCV}. The results are reported in Table~\ref{tab:dwarp}, where we can see a clear reduction of the error in character recognition that represents a $21\%$ performance increase.

\vspace{-0.25cm}
\subsection{Shadow Removal \& Non-Uniform Illumination.}
\begin{figure}[t]
    \centering
    \includegraphics[width=0.9\textwidth]{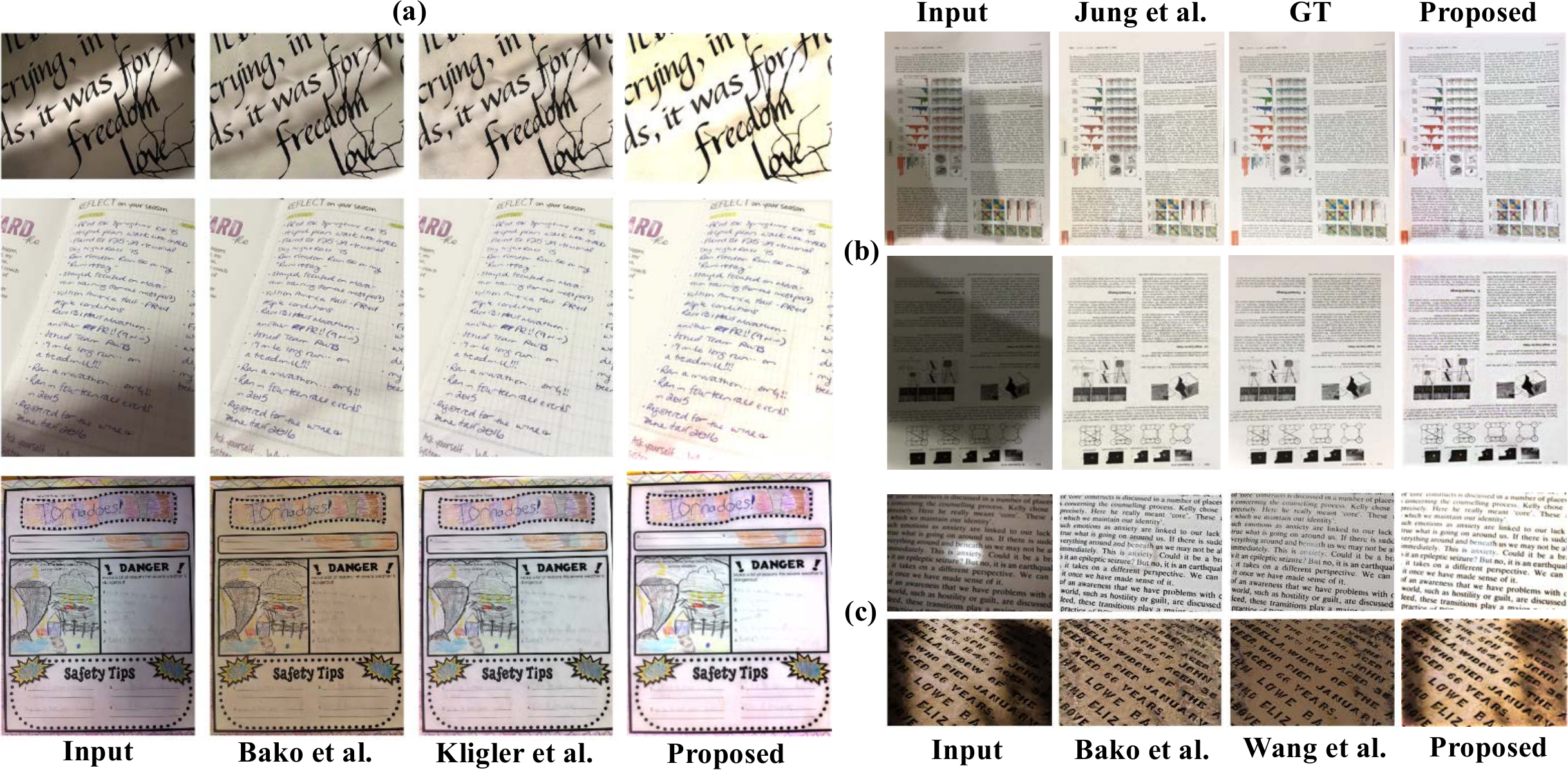}
    \vspace{3mm}
    \caption{Comparison with existing shadow removal methods \cite{bako2016removing,kligler2018document, jung2018water,wang2019effective} on real image sets: (a), (b), (c) are obtained from \cite{bako2016removing, jung2018water, wang2019effective} respectively. These comparisons show our method well generalizes on soft shadows. We report a fail case on hard shadows at the bottom row of (c).}
    \label{fig:shadow_comp}
    \vspace{-4mm}
\end{figure}

\begin{figure}[h!]
    \centering
    \includegraphics[width=0.9\textwidth]{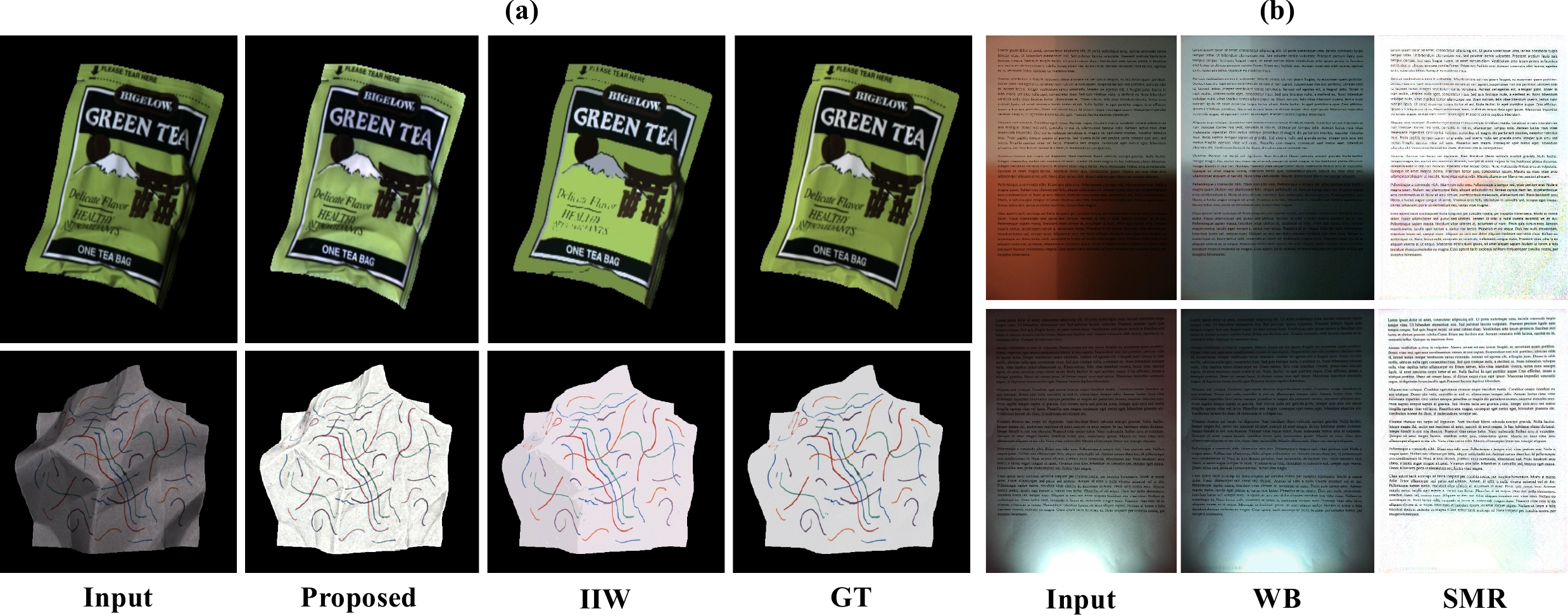}
    \vspace{3mm}
    \caption{(a) Comparison with IIW method \cite{bell14intrinsic}, IIW fails to accurately preserve text. (b) Results on multi-illuminant OCR dataset~\cite{michalak2020robust}, WB is white-balanced image and SMR is output after removing material and shading.}
    \label{fig:mitwetzu_comp}
    \vspace{-4mm}
\end{figure}

Although our method is not explicitly trained for shadow removal tasks it shows competitive performance when compared to previous shadow removal approaches \cite{bako2016removing,kligler2018document, jung2018water,wang2019effective}. We show the qualitative comparison in figure, \ref{fig:shadow_comp}(a), \ref{fig:shadow_comp}(b) and \ref{fig:shadow_comp}(c) on the real benchmark datasets provided in the works \cite{bako2016removing}, \cite{jung2018water}, and \cite{wang2019effective} respectively. Our method consistently performs well on different examples with soft shadows and shading but fails on hard shadow cases such as the image at the bottom row of \ref{fig:shadow_comp}(c). Additionally, in figure \ref{fig:mitwetzu_comp}(b) we show the performance of our method in challenging multi-illuminant conditions with shadows and shadings available in a recent OCR dataset~\cite{michalak2020robust}. We can see the results of after our white-balancing (WBNet) in the WB column of \ref{fig:mitwetzu_comp}(b), and in column SMR we can see how SMTNet gracefully handles strong illumination  conditions in real scenarios.
\vspace{-3mm}

\vspace{-4mm}
\section{Conclusions}
\vspace{-2mm}
In this work, we present  a unified learning-based architecture that directly estimates document reflectance based on intrinsic image formation. We achieve compelling results in intrinsic image decomposition of documents and document shading removal under challenging non-uniform illumination conditions. Additionally, we contribute a large multi-illuminant document dataset that extends the public dataset Doc3D~\cite{Das_2019_ICCV}.
The primary limitation of our model is the absence of explicit modeling for shadows. Thus, Doc3DShade does not have many hard shadow examples and the few shadow instances are self-shadows of document shapes. Therefore our proposed model does not generalize well for arbitrary hard shadows. In future work, we will explicitly model shadows in our self-supervised architecture. 

\noindent \textbf{Acknowledgements:}
This work has been partially supported by project TIN2014-61068-R, FPI Predoctoral Grant (BES-2015-073722), project RTI2018-095645-B-C21 of Spanish Ministry of Economy,  Competitiveness and the CERCA Programme / Generalitat de Catalunya, SAMSUNG Global Research Outreach (GRO) Program, NSF grants CNS-1718014, IIS-1763981, a gift from the Nvidia Corporation, a gift from Adobe Research, the Partner University Fund and the SUNY2020 ITSC.

\bibliography{egbib}

\begin{thebibliography}{52}
\providecommand{\natexlab}[1]{#1}
\providecommand{\url}[1]{\texttt{#1}}
\expandafter\ifx\csname urlstyle\endcsname\relax
  \providecommand{\doi}[1]{doi: #1}\else
  \providecommand{\doi}{doi: \begingroup \urlstyle{rm}\Url}\fi

\bibitem[Bako et~al.(2016)Bako, Darabi, Shechtman, Wang, Sunkavalli, and
  Sen]{bako2016removing}
Steve Bako, Soheil Darabi, Eli Shechtman, Jue Wang, Kalyan Sunkavalli, and
  Pradeep Sen.
\newblock Removing shadows from images of documents.
\newblock In \emph{Asian Conference on Computer Vision}, pages 173--183.
  Springer, 2016.

\bibitem[Barron and Malik(2015)]{Barron2015}
Jonathan~T. Barron and Jitendra Malik.
\newblock Shape, illumination, and reflectance from shading.
\newblock \emph{IEEE Transactions on Pattern Analysis and Machine
  Intelligence}, 37\penalty0 (8):\penalty0 1670--1687, 2015.

\bibitem[Barrow and Tenenbaum(1978)]{barrow1978recovering}
Harry Barrow and J~Tenenbaum.
\newblock Recovering intrinsic scene characteristics.
\newblock \emph{Comput. Vis. Syst}, 2, 1978.

\bibitem[Baslamisli et~al.(2018)Baslamisli, Le, and Gevers]{Baslamisli18}
Anil~S. Baslamisli, Hoang{-}An Le, and Theo Gevers.
\newblock {CNN} based learning using reflection and retinex models for
  intrinsic image decomposition.
\newblock In \emph{Computer Vision and Pattern Recognition}, 2018.

\bibitem[Bell et~al.(2014)Bell, Bala, and Snavely]{bell14intrinsic}
Sean Bell, Kavita Bala, and Noah Snavely.
\newblock Intrinsic images in the wild.
\newblock \emph{ACM Trans. on Graphics (SIGGRAPH)}, 33\penalty0 (4), 2014.

\bibitem[Bianco et~al.(2015)Bianco, Cusano, and Schettini]{bianco2015color}
Simone Bianco, Claudio Cusano, and Raimondo Schettini.
\newblock Color constancy using cnns.
\newblock In \emph{Proceedings of the IEEE Conference on Computer Vision and
  Pattern Recognition Workshops}, pages 81--89, 2015.

\bibitem[Brown and Tsoi(2006)]{brown2006geometric}
Michael~S Brown and Y-C Tsoi.
\newblock Geometric and shading correction for images of printed materials
  using boundary.
\newblock \emph{IEEE Transactions on Image Processing}, 15\penalty0
  (6):\penalty0 1544--1554, 2006.

\bibitem[Community()]{blender_2018}
Blender~Online Community.
\newblock Blender - a {{3D}} modelling and rendering package.
\newblock \url{www.blender.org}.

\bibitem[Das et~al.(2019)Das, Ma, Shu, Samaras, and Shilkrot]{Das_2019_ICCV}
Sagnik Das, Ke~Ma, Zhixin Shu, Dimitris Samaras, and Roy Shilkrot.
\newblock Dewarpnet: Single-image document unwarping with stacked 3d and 2d
  regression networks.
\newblock In \emph{The IEEE International Conference on Computer Vision
  (ICCV)}, October 2019.

\bibitem[Doersch et~al.(2015)Doersch, Gupta, and
  Efros]{doersch2015unsupervised}
Carl Doersch, Abhinav Gupta, and Alexei~A Efros.
\newblock Unsupervised visual representation learning by context prediction.
\newblock In \emph{Proceedings of the IEEE international conference on computer
  vision}, pages 1422--1430, 2015.

\bibitem[Fan et~al.(2018)Fan, Yang, Hua, Chen, and Wipf]{Fan2018RevisitingDI}
Qingnan Fan, Jiaolong Yang, Gang Hua, Baoquan Chen, and David~P. Wipf.
\newblock Revisiting deep intrinsic image decompositions.
\newblock \emph{2018 IEEE/CVF Conference on Computer Vision and Pattern
  Recognition}, pages 8944--8952, 2018.

\bibitem[Funt et~al.(1992)Funt, Drew, and Brockington]{Funt1992}
Brian Funt, Mark Drew, and Michael Brockington.
\newblock Recovering shading from color images.
\newblock In \emph{European Conference on Computer Vision}, pages 124--132,
  1992.

\bibitem[Gehler et~al.(2011)Gehler, Rother, Kiefel, Zhang, and
  Sch{\"o}lkopf]{Gehler2011}
Peter~V. Gehler, Carsten Rother, Martin Kiefel, Lumin Zhang, and Bernhard
  Sch{\"o}lkopf.
\newblock Recovering intrinsic images with a global sparsity prior on
  reflectance.
\newblock In \emph{Neural Information Processing Systems}, pages 765--773,
  2011.

\bibitem[Gijsenij and Gevers(2010)]{gijsenij2010color}
Arjan Gijsenij and Theo Gevers.
\newblock Color constancy using natural image statistics and scene semantics.
\newblock \emph{IEEE Transactions on Pattern Analysis and Machine
  Intelligence}, 33\penalty0 (4):\penalty0 687--698, 2010.

\bibitem[Gulrajani et~al.(2017)Gulrajani, Ahmed, Arjovsky, Dumoulin, and
  Courville]{gulrajani2017improved}
Ishaan Gulrajani, Faruk Ahmed, Martin Arjovsky, Vincent Dumoulin, and Aaron~C
  Courville.
\newblock Improved training of wasserstein gans.
\newblock In \emph{Advances in neural information processing systems}, pages
  5767--5777, 2017.

\bibitem[Haefner et~al.(2018)Haefner, Quéau, Möllenhoff, and
  Cremers]{Haefner2018CVPR}
B.~Haefner, Y.~Quéau, T.~Möllenhoff, and D.~Cremers.
\newblock Fight ill-posedness with ill-posedness: Single-shot variational depth
  super-resolution from shading.
\newblock In \emph{IEEE Conference on Computer Vision and Pattern Recognition
  (CVPR)}, 2018.

\bibitem[Horn(1974)]{horn1974determining}
Berthold~KP Horn.
\newblock Determining lightness from an image.
\newblock \emph{Computer graphics and image processing}, 3\penalty0
  (4):\penalty0 277--299, 1974.

\bibitem[Hu et~al.(2017)Hu, Wang, and Lin]{hu2017fc4}
Yuanming Hu, Baoyuan Wang, and Stephen Lin.
\newblock Fc4: Fully convolutional color constancy with confidence-weighted
  pooling.
\newblock In \emph{Proceedings of the IEEE Conference on Computer Vision and
  Pattern Recognition}, pages 4085--4094, 2017.

\bibitem[Hui et~al.(2016{\natexlab{a}})Hui, Sankaranarayanan, Sunkavalli, and
  Hadap]{hui2016white}
Zhuo Hui, Aswin~C Sankaranarayanan, Kalyan Sunkavalli, and Sunil Hadap.
\newblock White balance under mixed illumination using flash photography.
\newblock In \emph{2016 IEEE International Conference on Computational
  Photography (ICCP)}, pages 1--10. IEEE, 2016{\natexlab{a}}.

\bibitem[Hui et~al.(2016{\natexlab{b}})Hui, Sankaranarayanan, Sunkavalli, and
  Hadap]{iccp16wp}
Zhuo Hui, Aswin~C. Sankaranarayanan, Kalyan Sunkavalli, and Sunil Hadap.
\newblock White balance under mixed illumination using flash photography.
\newblock In \emph{IEEE Intl. Conf. Computational Photography (ICCP)},
  2016{\natexlab{b}}.

\bibitem[Hui et~al.(2019)Hui, Chakrabarti, Sunkavalli, and
  Sankaranarayanan]{hui2019learning}
Zhuo Hui, Ayan Chakrabarti, Kalyan Sunkavalli, and Aswin~C Sankaranarayanan.
\newblock Learning to separate multiple illuminants in a single image.
\newblock In \emph{Proceedings of the IEEE Conference on Computer Vision and
  Pattern Recognition}, pages 3780--3789, 2019.

\bibitem[Isola et~al.(2017)Isola, Zhu, Zhou, and Efros]{isola2017image}
Phillip Isola, Jun-Yan Zhu, Tinghui Zhou, and Alexei~A Efros.
\newblock Image-to-image translation with conditional adversarial networks.
\newblock In \emph{Proceedings of the IEEE conference on computer vision and
  pattern recognition}, pages 1125--1134, 2017.

\bibitem[Janner et~al.(2017)Janner, Wu, Kulkarni, Yildirim, and
  Tenenbaum]{janner2017self}
Michael Janner, Jiajun Wu, Tejas~D Kulkarni, Ilker Yildirim, and Josh
  Tenenbaum.
\newblock Self-supervised intrinsic image decomposition.
\newblock In \emph{Advances in Neural Information Processing Systems}, pages
  5936--5946, 2017.

\bibitem[Jung et~al.(2018)Jung, Hasan, and Kim]{jung2018water}
Seungjun Jung, Muhammad~Abul Hasan, and Changick Kim.
\newblock Water-filling: An efficient algorithm for digitized document shadow
  removal.
\newblock In \emph{Asian Conference on Computer Vision}, pages 398--414.
  Springer, 2018.

\bibitem[Kim et~al.(2016)Kim, Park, Sohn, and Lin]{Kim2016UnifiedDP}
Seungryong Kim, Kihong Park, Kwanghoon Sohn, and Stephen Lin.
\newblock Unified depth prediction and intrinsic image decomposition from a
  single image via joint convolutional neural fields.
\newblock In \emph{ECCV}, 2016.

\bibitem[Kligler et~al.(2018)Kligler, Katz, and Tal]{kligler2018document}
Netanel Kligler, Sagi Katz, and Ayellet Tal.
\newblock Document enhancement using visibility detection.
\newblock In \emph{Proceedings of the IEEE Conference on Computer Vision and
  Pattern Recognition}, pages 2374--2382, 2018.

\bibitem[Land and McCann(1971)]{Land1971}
Edwin~H. Land and John McCann.
\newblock Lightness and retinex theory.
\newblock \emph{Journal of the Optical Society of America}, 61\penalty0
  (1):\penalty0 1--11, 1971.

\bibitem[Lettry et~al.(2018)Lettry, Vanhoey, and {Van Gool}]{LVvG18}
Louis Lettry, Kenneth Vanhoey, and Luc {Van Gool}.
\newblock {Unsupervised Deep Single-Image Intrinsic Decomposition using
  Illumination-Varying Image Sequences}.
\newblock \emph{Computer Graphics Forum (Proceedings of Pacific Graphics)},
  37\penalty0 (10), October 2018.

\bibitem[Li and Snavely(2018)]{ZhenggiCVPR2018Learning}
Zhengqi Li and Noah Snavely.
\newblock Learning intrinsic image decomposition from watching the world.
\newblock In \emph{2018 CVPR}, pages 9039--9048, 2018.

\bibitem[Lou et~al.(2015)Lou, Gevers, Hu, Lucassen, et~al.]{lou2015color}
Zhongyu Lou, Theo Gevers, Ninghang Hu, Marcel~P Lucassen, et~al.
\newblock Color constancy by deep learning.
\newblock In \emph{BMVC}, pages 76--1, 2015.

\bibitem[Ma et~al.(2018{\natexlab{a}})Ma, Shu, Bai, Wang, and
  Samaras]{ma2018docunet}
Ke~Ma, Zhixin Shu, Xue Bai, Jue Wang, and Dimitris Samaras.
\newblock Docunet: document image unwarping via a stacked u-net.
\newblock In \emph{Proceedings of the IEEE Conference on Computer Vision and
  Pattern Recognition}, pages 4700--4709, 2018{\natexlab{a}}.

\bibitem[Ma et~al.(2018{\natexlab{b}})Ma, Chu, Zhou, Urtasun, and
  Torralba]{Ma2018SingleII}
Wei-Chiu Ma, Hang Chu, Bolei Zhou, Raquel Urtasun, and Antonio Torralba.
\newblock Single image intrinsic decomposition without a single intrinsic
  image.
\newblock In \emph{ECCV}, 2018{\natexlab{b}}.

\bibitem[Michalak and Okarma(2020)]{michalak2020robust}
Hubert Michalak and Krzysztof Okarma.
\newblock Robust combined binarization method of non-uniformly illuminated
  document images for alphanumerical character recognition.
\newblock \emph{Sensors}, 20\penalty0 (10):\penalty0 2914, 2020.

\bibitem[Narihira et~al.(2015)Narihira, Maire, and Yu]{Narihira2015}
Takuya Narihira, Michael Maire, and Stella~X. Yu.
\newblock Direct intrinsics: Learning albedo-shading decomposition by
  convolutional regression.
\newblock In \emph{International Conference on Computer Vision (ICCV)}, 2015.

\bibitem[Nayak and Nayak(2014)]{nayak2014odia}
Mamata Nayak and Ajit~Kumar Nayak.
\newblock Odia characters recognition by training tesseract ocr engine.
\newblock \emph{International Journal of Computer Applications}, 975:\penalty0
  8887, 2014.

\bibitem[Nestmeyer and Gehler(2017)]{Nestmeyer2017ReflectanceAF}
Thomas Nestmeyer and Peter~V. Gehler.
\newblock Reflectance adaptive filtering improves intrinsic image estimation.
\newblock \emph{2017 IEEE Conference on Computer Vision and Pattern Recognition
  (CVPR)}, pages 1771--1780, 2017.

\bibitem[Ronneberger et~al.(2015)Ronneberger, Fischer, and
  Brox]{ronneberger2015u}
Olaf Ronneberger, Philipp Fischer, and Thomas Brox.
\newblock U-net: Convolutional networks for biomedical image segmentation.
\newblock In \emph{International Conference on Medical image computing and
  computer-assisted intervention}, pages 234--241. Springer, 2015.

\bibitem[Shen et~al.(2013)Shen, Yeo, and Hua]{LShen2013}
Li~Shen, Chuohao Yeo, and Binh-Son Hua.
\newblock Intrinsic image decomposition using a sparse representation of
  reflectance.
\newblock \emph{IEEE Transactions on Pattern Analysis and Machine
  Intelligence}, 35\penalty0 (12):\penalty0 2904--2915, 2013.

\bibitem[Shi et~al.(2016)Shi, Loy, and Tang]{shi2016deep}
Wu~Shi, Chen~Change Loy, and Xiaoou Tang.
\newblock Deep specialized network for illuminant estimation.
\newblock In \emph{European Conference on Computer Vision}, pages 371--387.
  Springer, 2016.

\bibitem[Sial et~al.(2020)Sial, Baldrich, and Vanrell]{Sial:20}
Hassan~A. Sial, Ramon Baldrich, and Maria Vanrell.
\newblock Deep intrinsic decomposition trained on surreal scenes yet with
  realistic light effects.
\newblock \emph{J. Opt. Soc. Am. A}, 37\penalty0 (1):\penalty0 1--15, Jan 2020.

\bibitem[Sun et~al.(2019)Sun, Barron, Tsai, Xu, Yu, Fyffe, Rhemann, Busch,
  Debevec, and Ramamoorthi]{sun2019single}
Tiancheng Sun, Jonathan~T Barron, Yun-Ta Tsai, Zexiang Xu, Xueming Yu, Graham
  Fyffe, Christoph Rhemann, Jay Busch, Paul Debevec, and Ravi Ramamoorthi.
\newblock Single image portrait relighting.
\newblock \emph{ACM Transactions on Graphics (Proceedings SIGGRAPH)}, 2019.

\bibitem[Tappen et~al.(2006)Tappen, Adelson, and Freeman]{Tappen2006}
Marshall~F. Tappen, Edward~H. Adelson, and William~T. Freeman.
\newblock Estimating intrinsic component images using non-linear regression.
\newblock In \emph{IEEE Conference on Computer Vision and Pattern Recognition},
  pages 1992--1999, 2006.

\bibitem[Van De~Weijer et~al.(2007)Van De~Weijer, Gevers, and
  Gijsenij]{van2007edge}
Joost Van De~Weijer, Theo Gevers, and Arjan Gijsenij.
\newblock Edge-based color constancy.
\newblock \emph{IEEE Transactions on image processing}, 16\penalty0
  (9):\penalty0 2207--2214, 2007.

\bibitem[Wang and Chen(2019)]{wang2019effective}
Bingshu Wang and CL~Philip Chen.
\newblock An effective background estimation method for shadows removal of
  document images.
\newblock In \emph{2019 IEEE International Conference on Image Processing
  (ICIP)}, pages 3611--3615. IEEE, 2019.

\bibitem[Wang et~al.(2003)Wang, Simoncelli, and Bovik]{wang2003multiscale}
Zhou Wang, Eero~P Simoncelli, and Alan~C Bovik.
\newblock Multiscale structural similarity for image quality assessment.
\newblock In \emph{The Thrity-Seventh Asilomar Conference on Signals, Systems
  \& Computers, 2003}, volume~2, pages 1398--1402. Ieee, 2003.

\bibitem[Wei et~al.(2018)Wei, Lim, Zisserman, and Freeman]{wei2018learning}
Donglai Wei, Joseph~J Lim, Andrew Zisserman, and William~T Freeman.
\newblock Learning and using the arrow of time.
\newblock In \emph{Proceedings of the IEEE Conference on Computer Vision and
  Pattern Recognition}, pages 8052--8060, 2018.

\bibitem[Weiss(2001)]{Weiss2001}
Yair Weiss.
\newblock Deriving intrinsic images from image sequences.
\newblock In \emph{International Conference on Computer Vision}, pages 68--75,
  2001.

\bibitem[Wyszecki and Stiles(2000)]{colorscience07}
Gunther Wyszecki and W.~Stiles.
\newblock Color science: Concepts and methods, quantitative data and formulae,
  2nd edition.
\newblock \emph{Color Research \& Application}, 07 2000.

\bibitem[You et~al.(2017)You, Matsushita, Sinha, Bou, and
  Ikeuchi]{you2017multiview}
Shaodi You, Yasuyuki Matsushita, Sudipta Sinha, Yusuke Bou, and Katsushi
  Ikeuchi.
\newblock Multiview rectification of folded documents.
\newblock \emph{IEEE transactions on pattern analysis and machine
  intelligence}, 40\penalty0 (2):\penalty0 505--511, 2017.

\bibitem[Zhang et~al.(2007)Zhang, Yip, and Tan]{zhang2007removing}
Li~Zhang, Andy~M Yip, and Chew~Lim Tan.
\newblock Removing shading distortions in camera-based document images using
  inpainting and surface fitting with radial basis functions.
\newblock In \emph{Ninth International Conference on Document Analysis and
  Recognition (ICDAR 2007)}, volume~2, pages 984--988. IEEE, 2007.

\bibitem[Zhang et~al.(2016)Zhang, Isola, and Efros]{zhang2016colorful}
Richard Zhang, Phillip Isola, and Alexei~A Efros.
\newblock Colorful image colorization.
\newblock In \emph{European conference on computer vision}, pages 649--666.
  Springer, 2016.

\bibitem[Zhou et~al.(2015)Zhou, Kr{\"a}henb{\"u}hl, and
  Efros]{Zhou2015LearningDR}
Tinghui Zhou, Philipp Kr{\"a}henb{\"u}hl, and Alexei~A. Efros.
\newblock Learning data-driven reflectance priors for intrinsic image
  decomposition.
\newblock \emph{2015 IEEE International Conference on Computer Vision (ICCV)},
  pages 3469--3477, 2015.

\end{thebibliography}
\end{document}